\documentclass[runningheads]{llncs}
\usepackage[T1]{fontenc}
\usepackage{microtype}
\usepackage{graphicx}
\usepackage{booktabs} % for professional tables
\usepackage[switch]{lineno}
\usepackage{amsmath}
\usepackage{mathtools}
\usepackage{array}
\usepackage{multirow}
\usepackage{multicol}
\usepackage{enumitem}
\usepackage{soul}
\usepackage[caption=false,font=normalsize,labelfont=sf,textfont=sf]{subfig}
\usepackage{url}
\usepackage{hyperref}
\usepackage{marvosym}
\usepackage[noabbrev]{cleveref}
\usepackage{amsfonts}
\usepackage{subfig}
\usepackage[autostyle]{csquotes}  
\usepackage{mwe}
\usepackage{physics}
\usepackage{tikz}
\begin{document}
\title{Toward Artificial General Intelligence via Intelligence Foundation Models}
\author{Borui Cai\inst{1} \and
Yao Zhao\inst{2\textrm{(\Letter)}}}
\institute{Hangzhou International Innovation Institute, Beihang University, China.
\email{caibr@buaa.edu.cn} \and
Victoria University, Melbourne, Australia\\
\email{yao.zhao@vu.edu.au}}
\maketitle              % typeset the header of the contribution
\begin{abstract}
We propose a new perspective for approaching artificial general intelligence (AGI) through an Intelligence Foundation Model (IFM).
Unlike existing foundation models (FMs), which learn statistical regularities within surface-level manifestations of human intelligence, such as language or vision, IFM aims to infer the underlying structural principles that give rise to human intelligence itself.
This perspective is grounded in the view that human intelligence arises from stable neural dynamics in the human brain and is expressed through diverse behaviors across environments.
Accordingly, IFM learns from a broad spectrum of human intelligent behaviors to recover the structural neural dynamics that generate intelligence.
For this perspective, IFM introduces two core components: a neural system architecture and a neuron output prediction objective. The architecture models the temporal evolution of biologically inspired neural dynamics, enabling the system to store, integrate, and process information over time. The neuron output prediction objective provides a unified computational principle for learning the underlying neural dynamics that give rise to diverse human intelligent behaviors.
Together, these components establish a biologically grounded and computationally feasible framework for developing systems with the potential for generalization, reasoning, and adaptive learning across domains, offering a promising step toward AGI.

\end{abstract}
\section{Introduction}
Artificial general intelligence (AGI) refers to systems capable of understanding, learning, and generating adaptive behavior across diverse tasks and environments, comparable to or exceeding human-level intelligence~\cite{agi}. Yet, despite decades of progress, realizing this level of generality remains an open challenge. To advance toward AGI, we propose a novel perspective through an \textbf{Intelligence Foundation Model (IFM)}. Rather than treating human intelligence as a collection of surface-level abilities (e.g., language, vision, or motor skills), IFM conceptualizes human intelligence as an emergent property of stable neural dynamics in the human brain, which is honed by millions of years of evolution and enables diverse adaptive behaviors in the physical world \cite{humanbrain}. To date, the human brain remains the only known system that instantiates general intelligence~\cite{humanagi}. While individual brains differ anatomically, they all realize a shared manifold of neural dynamics that supports general intelligent behavior across individuals. IFM therefore aims to \textit{learn the underlying structural neural dynamics that give rise to human intelligence}. Meanwhile, any intelligent behavior, whether perception, reasoning, planning, or decision-making, can ultimately be expressed as temporal transformations from inputs to outputs in the human brain. Therefore, understanding intelligence reduces to understanding the neural dynamics that give rise to such transformations. In this view, IFM reframes the challenge of ``learning intelligence” as the problem of capturing the structural neural dynamics governing such input-output transformations in the human brain. The shift from a task-driven paradigm to a structure-driven paradigm allows IFM to focus on the implicit principles that evolution has discovered for building general-purpose intelligent systems.
Inspired by the success of recent foundation models (FMs)~\cite{fm}, IFM seeks to leverage large-scale neuronal input-output transformation data that comprehensively captures diverse human behaviors to advance toward AGI.

Emerging work provides early evidence for this direction. Specifically, neural network models trained on large-scale neuronal recordings in animals such as mice and C. elegans have demonstrated the ability to predict both neural activity and behavior. In mice~\cite{mice}, models trained on large-scale visual cortex recordings can accurately predict population neural responses to naturalistic stimuli and generalize to novel, previously unseen stimulus types. 
Models trained on the full connectome of C. elegans have been shown to predict neuronal dynamics underlying locomotor behaviors, including crawling and nictation \cite{celegans}.
These results provide empirical support for the central premise of IFM, that neural dynamics underlying intelligent behavior can be learned from neural input–output data.

\begin{figure}[tp]
\centering
\includegraphics[width=0.6\linewidth]{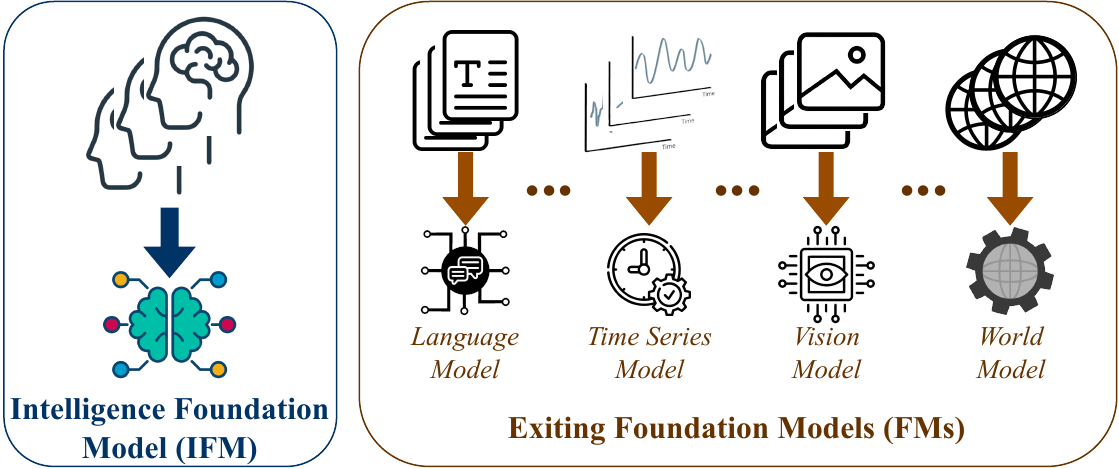}
\caption{Comparison between IFM and existing FMs.}
\label{fig:fig1_bkp}
\end{figure}

\vspace{0.1in}
\par \underline{\textbf{Fundamental barriers to AGI for existing FMs:}} 
Existing FMs adhere to the paradigm of large-scale, task-agnostic pretraining to learn universal representations that can be efficiently adapted to diverse downstream tasks and modalities \cite{fm1}. One of the examples is large language models (LLMs), e.g., ChatGPT \cite{chatgpt}, DeepSeek \cite{deepseek}, and Gemini \cite{gemini}, which approximate human ability in text processing and have demonstrated impressive capabilities across a wide range of natural language tasks, e.g., question answering \cite{question}, translation \cite{translation}, summarization \cite{summarization}, and code generation \cite{codegen}. However, existing FMs constantly face limitations across various intelligent capabilities, including but not limited to:

\begin{itemize}[leftmargin=*, noitemsep]
    \item \textit{Reasoning}: FMs can exhibit limitations in reasoning, occasionally generating factually incorrect or fabricated content with high confidence \cite{reasoning}.
    \item \textit{Learning:} FMs often face challenges in adapting behavior based on new experiences, typically relying on retraining or fine-tuning \cite{lora}.
    \item \textit{Memory:} FMs have limited capacity to autonomously acquire and retain new knowledge, often depending on external databases or retrieval systems \cite{memory}.
\end{itemize}

A core reason underlying these limitations is that existing FMs primarily learn patterns within specific domains, rather than capturing the broader structure of intelligence~\cite{embers}. 
For example, LLMs are difficult at learning because language alone serves as a descriptive medium rather than a functional substrate for learning. While linguistic representations can articulate the principles of learning, they cannot embody the complex neural dynamics through which biological learning occurs.
A straightforward comparison is illustrated in Fig.~\ref{fig:fig1_bkp}: LLMs capture statistical regularities in human language; time series FMs model temporal dependencies in domains such as financial~\cite{financial} and weather data~\cite{weather}; vision FMs encode spatial, temporal, and semantic continuity in images~\cite{image} and videos~\cite{video}; and world models~\cite{world} learn the dynamics and causal relationships of physical environments. Recent multimodal FMs \cite{multimodal} focus on aligning and integrating information across data modalities, but largely operate by capturing cross-modal statistical regularities. 
Overall, existing FMs capture only partial expressions or surface regularities of intelligence, but are not designed to model its underlying principles comprehensively~\cite{tool}.

\vspace{0.1in}
\par \underline{\textbf{Key advantages of IFM for AGI:}} 
Unlike existing FMs, which are designed for specific modalities such as language or vision, IFM learns neural dynamics underlying human intelligence directly from diverse human behavior that embodies broad intelligence.
However, intelligence has long been an elusive subject, and numerous theoretical frameworks have attempted to explain its nature. Predictive processing theories, including Free Energy Principle (FEP) \cite{fep} and the predictive mind hypothesis \cite{predict}, conceptualize the brain as a predictive machine that seeks to minimize the discrepancy between internal predictions and sensory observations. Information-integration theories, such as Global Workspace Theory (GWT) \cite{gwt1}, emphasize that intelligence emerges from the integration of specialized modules that share and broadcast information across a global neural workspace \cite{gwt2}. Other approaches focus on more localized mechanisms, such as the complementary learning systems, which highlight the distinct yet cooperative roles of the hippocampus and neocortex in forming long- and short-term memories \cite{hippo}, or studies that track how the hippocampus and striatum encode value signals during decision-making \cite{hippo2}.
While these theories offer valuable insights into aspects of cognition, they primarily explain why human intelligence arises rather than how to model and replicate it in machines at scale for building AGI.

Human intelligence emerges from the collective dynamics of roughly 86 billion neurons~\cite{86}. However, the immense complexity of these interactions makes it extremely difficult to construct an explicit, mechanistic model of intelligence from first principles. To address this challenge, IFM is designed to capture the implicit structural neuronal dynamics that underlie observed intelligent behavior, learning directly from large-scale neuronal data representing diverse human cognitive processes. By grounding itself in observable behavior rather than abstract theoretical formulations, IFM avoids reliance on hand-crafted models of intelligence and offers a more practical pathway toward AGI.

\begin{figure}[tp]
\centering
\includegraphics[width=0.6\linewidth]{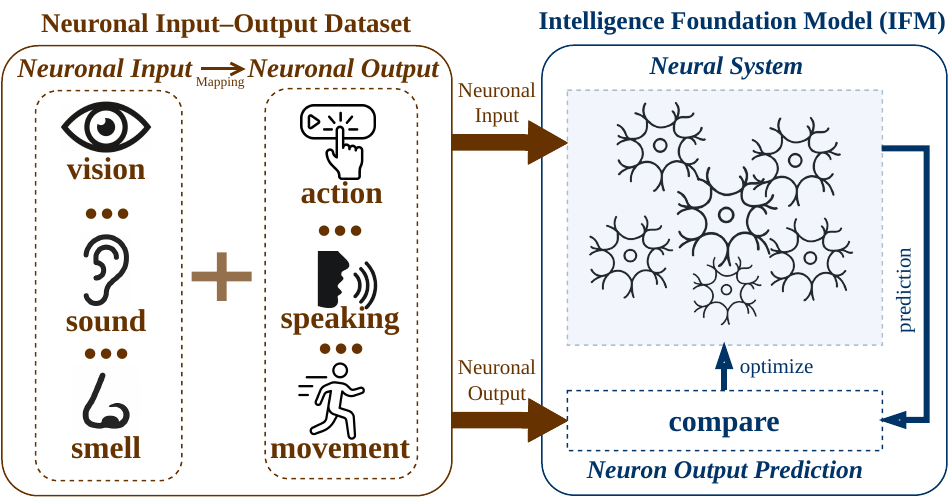}
\caption{Conceptual overview of IFM.}
\label{fig:fig2_bkp}
\end{figure}

\vspace{0.1in}
\par \underline{\textbf{Conceptual Overview of IFM:}}
As illustrated in Fig.~\ref{fig:fig2_bkp}, IFM learns the implicit structural neuronal dynamics that underlie observed intelligent behavior. It is expected to achieve this through two core designs:

\begin{itemize}[leftmargin=*, noitemsep]
    \item \emph{Neural System Architecture}: IFM is constructed upon a neural system architecture designed to model three key dynamic behaviors of biological neurons. First, each neuron maintains an internal state that encodes past activity to guide future decisions (\emph{Neuron Function}). Second, the neurons are interconnected through recurrent, graph-like connections that allow information to branch, merge, and loop across multiple routes, supporting a wide spectrum of intelligent functions (\emph{Neuron Connectivity}). Third,  connection strengths are allowed to evolve based on neuronal dynamics, allowing the network to self-organize, adapt, and continuously refine its internal structure through experience (\emph{Neuronal Plasticity}).
    \item \emph{Neuron Output Prediction}: The collective dynamics underlying diverse human intelligent behavior are formalized as generating neuronal outputs from the temporal input signals of biological neural systems. 
    Building on this formulation, IFM defines its training objective as predicting biological neuronal outputs from the corresponding input neuronal dynamics over time, thereby enabling the model to internalize the essential computational processes that underpin intelligence.
    This objective directly links low-level neuronal dynamics to high-level intelligent behavior, enabling the model to learn temporal, recurrent, and context-dependent transformations that characterize human intelligence. In essence, it bridges neural dynamics with the principles of intelligence.
    % \item \textbf{Intelligence Dataset Curation}. 
\end{itemize}

Building on these designs, IFM learns intelligence from structured datasets that encode intelligent behavior as temporal sequences of neuronal inputs and outputs, allowing training through standard backpropagation.

\section{IFM Technical Details}
IFM models the structural neuronal dynamics that underlie observed intelligent behavior. Its design is detailed as follows.

\subsection{Neural System Architecture}
We define the neural system for IFM as $\mathcal{G}=(\mathcal{N},\mathcal{E})$, where $\mathcal{N}=\{\eta_{1},\eta_{2},...,\eta_{n}\}$ denotes the neurons in $\mathcal{G}$, and $\mathcal{E}$ denotes the edges that connect these neurons. We categorize neurons as \textbf{input neurons}, \textbf{output neurons}, and \textbf{hidden neurons} as follows.

\begin{figure}[tp]
\centering
\includegraphics[width=0.6\linewidth]{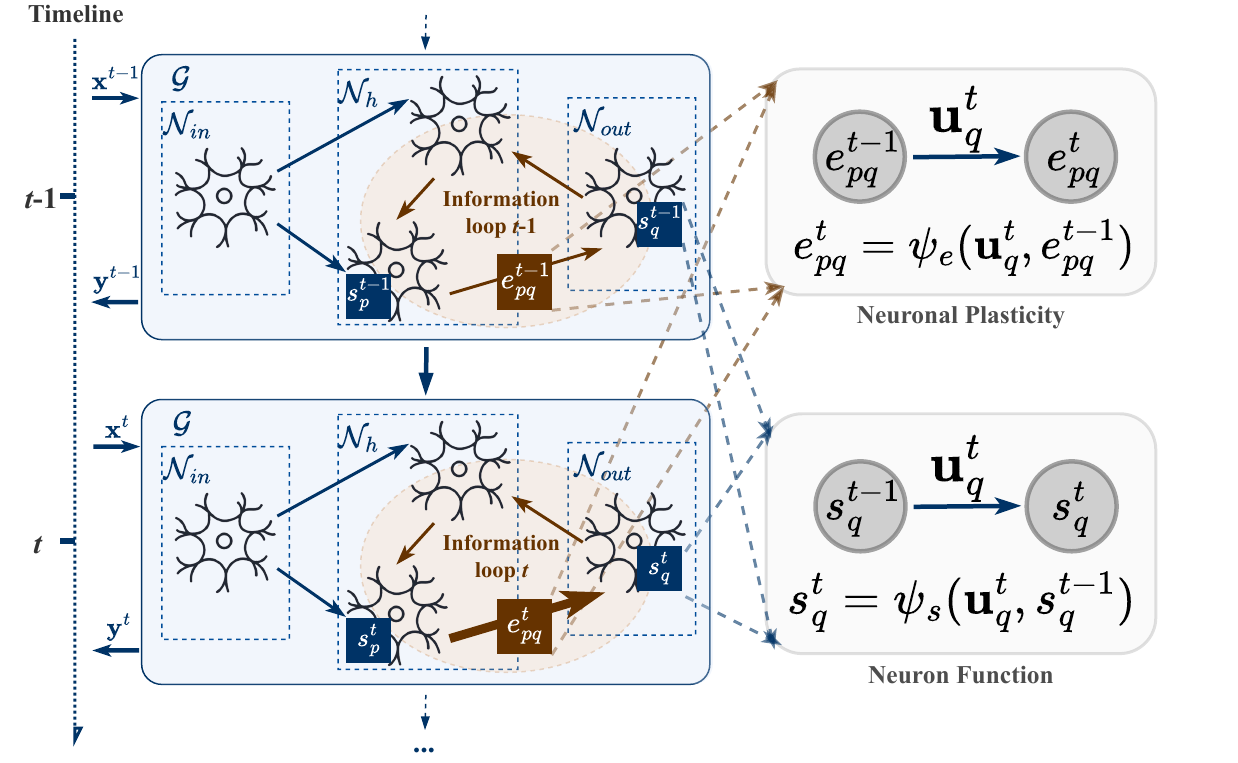}
\caption{Neural system of IFM.}
\label{fig:fig3_bkp}
\end{figure}

\begin{itemize}[leftmargin=*, noitemsep]
\item \emph{Input neurons} $\mathcal{N}_{in}\in \mathcal{N}$ are neurons that can receive stimulus signals originating from outside the system, analogous to human sensory neurons responsible for processing visual, auditory, olfactory, and other external inputs.
\item \emph{Output neurons} $\mathcal{N}_{out}\in \mathcal{N}$ are neurons whose activity can be externally perceived. This typically includes neurons that drive observable actions—such as motor neurons that control muscle contraction, and certain neurons whose activity reflects internal states of $\mathcal{G}$, providing insight into the network’s internal dynamics.
\item \emph{Hidden neurons} $\mathcal{N}_{h} \in \mathcal{N}$ are neurons that receive signals only from within the network and whose activity does not directly produce observable outputs. They process and integrate these signals, maintain the network’s internal state, and mediate information flow between input and output neurons, enabling the system to perform complex computations.
\end{itemize}

With these neurons, IFM neural system $\mathcal{G}$ receives an input stimulus sequence with input neurons $(\mathbf{x}^{1},\mathbf{x}^{2},…,\mathbf{x}^{T}),\ \forall \mathbf{x}^{t}\in \mathbb{R}^{|\mathcal{N}_{in}|}$ over time, and generates an output response sequence through output neurons $(\mathbf{y}^{1},\mathbf{y}^{2},…,\mathbf{y}^{T}),\ \forall \mathbf{y}^{t}\in \mathbb{R}^{|\mathcal{N}_{out}|}$, where $T$ is the timestamp. By that, the function of $\mathcal{G}$ can be defined as $\mathbf{y}^{t}=\mathcal{G}(\mathbf{x}^{1},\mathbf{x}^{2},...,\mathbf{x}^{t}),\ \forall 1\leq t\leq T$.

The design of the neural system $\mathcal{G}$ models the neuron dynamics of biological neural systems, based on three key components: \textbf{neuron function}, \textbf{neuron connectivity}, and \textbf{neuronal plasticity}. Neuron function captures the internal state and temporal evolution of biological neurons, enabling each neuron to integrate inputs and maintain memory over time. Neuron connectivity models the flexible, recurrent wiring of neurons in the brain, enabling signals to propagate through complex recurrent graphs that support oscillations, synchronization, and context-dependent computation. Neuronal plasticity captures the adaptive modulation of connection strengths in response to neuronal dynamics, reflecting biological learning mechanisms. Together, these three components establish a computational framework that embodies the dynamic, adaptive, and self-organizing nature of intelligence, as shown in Fig.~\ref{fig:fig3_bkp}.

\vspace{0.1in}
\par \underline{\textbf{Neuron Function:}} The function of a neuron $\eta \in \mathcal{N}$ is to maintain a dynamic internal state $s$ that integrates past information and transforms incoming signals into outputs over time, thereby enabling temporal and context-dependent behavior. This process can be formalized as $v^{t} = \phi(\mathbf{u}^{t}, s^{t-1})$, where $\mathbf{u}^{t}$ and $v^{t}$ denote the input and output at time $t$, respectively. In this formulation, as shown in Fig.~\ref{fig:fig3_bkp}, the neuron state $s^{t} = \psi_s(\mathbf{u}^{t}, s^{t-1})$ recursively integrates information from all previous time steps up to $t$. Recurrent~\cite{recurrent} and spiking neurons~\cite{spiking} can be regarded as two special cases within this general structure. Specifically, a recurrent neuron produces continuous outputs, with its internal state updated through a nonlinear transformation $s^{t} = \sigma(W_{\text{in}}\mathbf{u}^{t} + W_{s}s^{t-1})$, where $W_{\text{in}}$ and $W_{s}$ are the input and recurrent weight matrices, respectively. In contrast, a spiking neuron generates discrete spikes, with its internal state governed by membrane potential dynamics. For example, in the Leaky Integrate-and-Fire (LIF) model~\cite{lif}, the membrane potential evolves as $\frac{ds(t)}{dt} = -(s(t) - s_{\text{rest}}) + \mathbf{w}^{\top}\mathbf{u}(t)$, where $s_{\text{rest}}$ is the resting potential and $\mathbf{w}$ denotes the synaptic weights for each input.

\vspace{0.1in}
\par \underline{\textbf{Neuron Connectivity:}} 
In the neuron system $\mathcal{G}$, neurons are interconnected through directed, weighted edges represented by an adjacency matrix $\mathcal{E} \in \mathbb{R}^{\mathcal{N} \times \mathcal{N}}$, where $e_{pq}$ denotes the weight of the edge from neuron $\eta_p$ to neuron $\eta_q$. Unlike the layer-wise connectivity typically adopted in existing FMs, our approach employs a graph-structured neural system. This flexible connectivity supports both feedforward pathways and recurrent loops, allowing information to branch, merge, and circulate throughout the network. This allows signals to interact dynamically across time and pathways, enabling the model to capture temporal dependencies, integrate heterogeneous information, and coordinate complex behavior. These are capabilities that traditional layer-wise architectures inherently struggle to achieve.

Notably, recurrent connectivity is essential for modeling human intelligence but is largely overlooked in existing FMs. 
It enables information to persist and interact over time, creating temporal continuity that is critical for integrating past experiences, forming context-dependent representations, and coordinating complex behavior.
At the microcircuit level, two reciprocally connected neurons can form an oscillatory loop generating periodic signals, akin to pacemaker neurons whose intrinsic rhythms coordinate brain-wide activity~\cite{pacemaker}. 
Scaling up, larger neuronal assemblies sustain rhythmic firing patterns that synchronize activity across populations to support higher-order cognition~\cite{rhythms}.
Such dynamics mirror the neural oscillatory activity, such as gamma and theta brain waves (generated by excitatory–inhibitory loops in the cortex and hippocampus) \cite{brainwave}.

\vspace{0.1in}
\par \underline{\textbf{Neuronal Plasticity:}} 
Neuronal plasticity is a fundamental mechanism in biological neurons, supporting essential capabilities such as learning and memory \cite{hebb}. In biological neural systems, the strength and efficacy of synaptic connections can change over time in response to experience, sensory inputs, or environmental demands. This adaptability allows neural circuits to encode information, form memories, and reorganize themselves to optimize performance \cite{ltp}. 

Considering two neurons in neural system $\mathcal{G}$, $\eta_{p}$ and $\eta_{q}$, that are connected by an edge of weight $e_{pq}$, we define $\eta_{p}$ as the source (presynaptic) neuron that transmits a signal, and $\eta_{q}$ as the target (postsynaptic) neuron that receives it.
Neuronal plasticity is realized as the modulation of edge weights based on the historical activity of connected neurons.
That is, the weight $e_{pq}$ is influenced by the
outputs of both neurons, i.e., $v_{p}$ of $\eta_{p}$ (or $u_{q}$ of $\eta_{q}$, equivalently) and $v_{q}$ of $\eta_{q}$. Then, we have $e_{pq}^{t}=\psi_{e}(u_{q}^{1},...,u_{q}^{t},v_{q}^{1},...,v_{q}^{t})$.
Substituting $v_{q}=\phi(u_{q},s_{q})$, the update can be expressed recursively as $e_{pq}^{t}=\psi_{e}(u_{q}^{t},e_{pq}^{t-1})$, where $e_{pq}^{t-1}$ as the previous edge weight. This formulation naturally extends to target neurons with multiple inputs, i.e., $e_{pq}^{t}=\psi_{e}(\mathbf{u}_{q}^{t},e_{pq}^{t-1})$.

Many existing neural plasticity models, like Hebbian plasticity \cite{hebbian} and Spiking-Timing-Dependent-Plasticity (STDP) \cite{stdp} can be expressed in this framework. STDP can be formulated as $e_{pq}^{t} = \psi_e(\mathbf{u}_q^t, e_{pq}^{t-1}) = e_{pq}^{t-1} + \Delta e_{pq}^t$, where $\Delta e_{pq}^t$ is given by:
\begin{itemize}[leftmargin=*, noitemsep]
    \item \emph{Long-term Potentiation}: If $\eta_{p}$ fires shortly before $\eta_{q}$ (within a certain time window), $\Delta e_{pq}^t>0$.
    \item \emph{Long-term Depression}: If $\eta_{p}$ fires shortly after $\eta_{q}$ (within a certain time window), $\Delta e_{pq}^t<0$.
\end{itemize}

\subsection{Neuron Output Prediction}
Every manifestation of intelligent behavior, whether it involves pattern recognition, memory formation, causal reasoning, or decision-making, emerges from the transformation of neuronal inputs into outputs through complex and adaptive neural dynamics. Based on this principle, we introduce a neuron output prediction objective for IFM to learn from intelligent behavior represented as neuronal input–output sequences. This objective unifies diverse intelligent behaviors within a single learning framework and enables IFM to capture the underlying principles of intelligence.

\vspace{0.1in}
\par \underline{\textbf{Formal Definition:}} 
With ground truth input stimulus sequence $(\mathbf{x}^{1},\mathbf{x}^{2},…,\mathbf{x}^{T})$ and the response sequence $(\mathbf{y}^{1},\mathbf{y}^{2},…,\mathbf{y}^{T})$ sampled from real-world human intelligent behavior, the formulation of the neuron output prediction is defined as $\mathcal{L}=\min_{\mathcal{G}}\sum_{t=1}^{T}\ell(\mathbf{y}^{t},\hat{\mathbf{y}}^{t})$, with $\hat{\mathbf{y}}^{t}=\mathcal{G}(\mathbf{x}^{t},(\mathbf{s}^{t-1},\mathbf{e}^{t-1}))$. In the objective, $\mathcal{L}$ represents the total loss accumulated over a sequence of $T$ time steps, and $\ell(\mathbf{y}^{t}, \hat{\mathbf{y}}^{t})$ measures the discrepancy between the real neuronal output $\mathbf{y}^{t}$ and predicted output by IFM $\hat{\mathbf{y}}^{t}$ at time $t$. By minimizing $\mathcal{L}$, IFM learns to predict correct neuronal outputs over time, thereby capturing the temporal patterns and dependencies underlying intelligent behavior.

\begin{figure}[t!]
\centering
\includegraphics[width=0.8\linewidth]{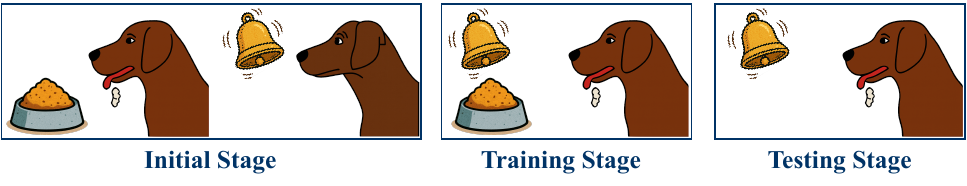}
\includegraphics[width=0.6\linewidth]{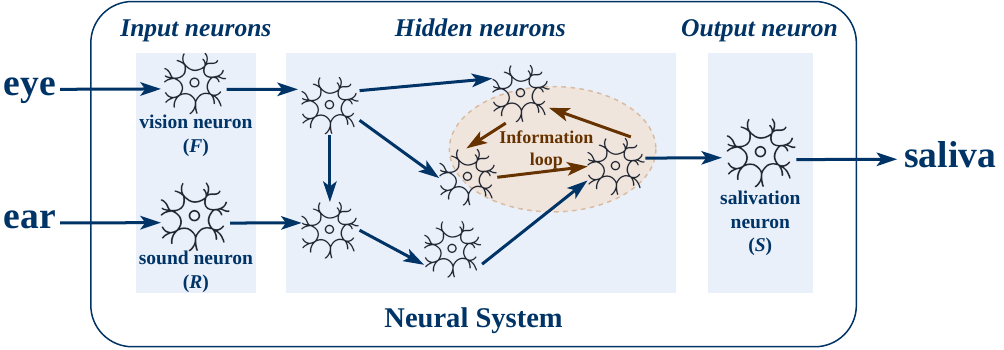}
\caption{Pavlov’s classical conditioning experiment.}
\label{fig:pavlov}
\end{figure}

\vspace{0.1in}
\par \underline{\textbf{Concept Illustration:}} 
We illustrate the neuron output prediction concept using Pavlov’s classical conditioning experiment \cite{pavlov}, which is a canonical example of biological learning behavior \cite{learning}. We first sample neuronal input–output sequences from the entire conditioning process, then show how predicting these neuronal outputs allows an IFM neural system to emulate this biological learning behavior.

As illustrated in Fig.~\ref{fig:pavlov}, the experiment shows that, through repeated training, a dog learns to associate the sound of a ringing bell with the presentation of food, eventually salivating at the sound alone. For simplicity, we represent the dog with three functional neurons: a vision neuron that detects food ($F$), a sound neuron that perceives ringing ($R$), and a salivation neuron that triggers salivation ($S$). The overall conditioning process has three stages:
\begin{itemize}[leftmargin=*, noitemsep]
\item \emph{Initial Stage}: The dog salivates when it sees food but not when it only hears the bell, i.e., $\{\mathbf{x}^{1}=(F,\neg R),\mathbf{x}^{2}=(\neg F,R)\}_{init},\{\mathbf{y}^{1}=S,\mathbf{y}^{2}=\neg S\}_{init}$.
\item \emph{Training Stage}: The dog simultaneously sees food and hears the bell over two trials, i.e., $\{\mathbf{x}^{3}=(F,R),\mathbf{x}^{4}=(F,R)\}_{train}$, $\{\mathbf{y}^{3}=S,\mathbf{y}^{4}=S\}_{train}$.
\item \emph{Testing Stage}: The dog has learned to associate the bell with food, i.e., 
$\{\mathbf{x}^{5}=(\neg F,R)\}_{test}$, $\{\mathbf{y}^{5}=S\}_{test}$. 
\end{itemize}

Therefore, the entire conditioning behavior can be represented as temporal sequences of neuronal inputs and outputs spanning the three stages, i.e., $\mathbf{x}=\{(F,\neg R),(\neg F,R),(F,R),(F,R),(\neg F,R)\}$ and $\mathbf{y}=\{S,\neg S,S,S,S\}$. Additional sequences can be generated by varying trial patterns, e.g., changing the number of trials per stage, to enrich the dataset and increase sample diversity.

Under the objective of neuron output prediction, an IFM neural system is trained to predict $\mathbf{y}$ from $\mathbf{x}$, demonstrating conditioning-like learning ability. Specifically, we train the neural system $\mathcal{G}$ using the above neuronal input–output sequences, where two input neurons receive $\mathbf{x}$ and one output neuron generates $\mathbf{y}$. The network includes multiple hidden neurons to learn the mapping from $\mathbf{x}$ to $\mathbf{y}$ by minimizing the neuron output prediction loss $\mathcal{L}$. Through this training process, $\mathcal{G}$ adjusts its internal parameters that regulate neuron function, neuron connectivity, and neuronal plasticity to emulate the conditioning mechanism and reproduce corresponding neuronal output patterns. Importantly, the network is not meant to memorize individual trials, but to internalize the fundamental principles of classical conditioning that generalize across many samples.

\section{IFM Training}
% \vspace{0.1in}
% \par \underline{\textbf{Intelligence Dataset Curation:}} 
% \subsection{Intelligence Dataset Curation} 
A major challenge in training IFM under the neuron output prediction objective is constructing neuronal input–output datasets that capture the temporal evolution of neural activities underlying intelligent behavior. Because IFM learns the time-dependent mapping between neuronal inputs and outputs, the training data should reflect the dynamic transformation of signals as the neural system interacts with its environment. Each data sample consists of a pair of neuronal input–output sequences, expressed as $d=\{(\mathbf{x}^{1},...,\mathbf{x}^{T}),(\mathbf{y}^{1},...,\mathbf{y}^{T})\}$, while the overall dataset comprises multiple such samples, denoted by $D=\{d_{1},d_{2},\ldots,d_{n}\}$. 
In line with the FM paradigm, data samples should be collected from a broad range of individuals performing diverse intelligent behaviors.
Such data samples can be obtained in two ways:

\begin{itemize}[leftmargin=*, noitemsep]
\item \emph{Direct Neuronal Sampling}: This approach acquires neuronal input–output data directly from biological brains by recording synaptic activity across neurons as subjects perform intelligent behavior. It provides the most biologically faithful supervision for IFM, as it reflects the genuine neuronal computations underlying cognition. However, large-scale data collection is technically infeasible, as capturing the full dynamics of even small cortical circuits at cellular resolution remains an open challenge. Current techniques, such as Neuropixels probes~\cite{Neuropixels} and calcium imaging~\cite{ultrasensitive}, can only record limited subsets of neurons and modalities, making this method suitable primarily for small-scale or proof-of-concept IFMs.

\item \emph{Indirect Neuronal Sampling}: This scalable alternative captures functional equivalents of neuronal inputs and outputs via fully instrumented, human-in-the-loop embodiments, such as neurosensory interfaces (wearable multimodal sensing systems) or avatars (digital or robotic agents with human-like perception and actuation). These embodiments perceive environmental stimuli (visual, auditory, tactile, proprioceptive) and record observable behavioral responses (speech, gestures, motor actions). The sensory streams serve as proxies for neuronal inputs, while behavioral responses represent neuronal outputs. Recording paired temporal sequences during natural interactions produces high-dimensional data encoding the dynamics of intelligent behavior. Training IFM on such data enables it to learn how sensory experiences evolve into actions over time, similar to imitation learning \cite{imitation}, effectively capturing neuron-like transformation principles without requiring direct observation of biological neurons.
\end{itemize}

With advances in neural recording technologies, direct and indirect neuronal sampling are expected to converge. This convergence will enable IFM to go beyond emulating intelligent behavior, moving toward internalizing its underlying principles and thus closing the loop between cognition and computation. Using these data samples, IFM implemented by differentiable neural systems can be efficiently trained via backpropagation. Given the streaming nature of the training data, variants of Truncated Backpropagation Through Time (TBPTT)~\cite{tbptt} can be employed to reduce computational complexity. 

% We present a \emph{toy example} in which IFM is used to train an artificial Pong player, demonstrating the overall design process available at \href{https://github.com/brcai/IFM_pong}{https://github.com/brcai/IFM\_pong}. 

% \section{Call to Action}
% {\color{red} XX}

\section{Discussion and Significance}
\par \underline{\textbf{Discussion:}} Despite variations in age, culture, and individual biology, human intelligence consistently arises from shared fundamental principles embedded in the brain. Building on this premise, IFM leverages the foundation model paradigm to capture these underlying mechanisms of neuronal dynamics in the human brain. This approach defines a new perspective toward AGI, one that models the principles of intelligence itself rather than its surface-level manifestations. We envisage a pragmatic and scalable roadmap toward AGI through IFM from two complementary trajectories (\textbf{Call to Action}): 
\begin{itemize}[leftmargin=*, noitemsep]
    \item \textit{Biological scaling} begins with simpler neural systems or lower organisms and progressively advances to more complex brains, systematically revealing the principles of intelligence across increasing levels of complexity.
    \item \textit{Functional scaling} develops IFMs for specific robotic applications, such as industrial manipulators or household assistants, and gradually integrates these specialized capabilities into a unified, general-purpose intelligent system.
\end{itemize}

Despite its promise, realizing this vision faces several major challenges. Acquiring high-resolution, longitudinal neural activity data across diverse species and brain regions is technically demanding. Designing neural systems that faithfully capture recurrent, plastic, and temporally extended neural dynamics while remaining computationally tractable presents significant modeling challenges. Finally, training IFM to reproduce complex neural dynamics across multiple systems and scales requires novel optimization strategies to ensure convergence and generalization.
Overcoming these challenges will be essential to advancing IFM as a practically viable approach toward general-purpose intelligence.

\par \underline{\textbf{Significance:}} By reframing intelligence as the product of underlying structural neural dynamics rather than isolated surface-level manifestations, IFM provides a paradigm shift with broad implications for both neuroscience and AI. This perspective suggests that AGI may not require reproducing the full biological intricacies of the brain, but instead capturing the dynamical principles that allow artificial neural systems to generate diverse intelligent behavior. 
IFM thus offers a framework that potentially bridges three traditionally separate areas: neural circuit dynamics, models of learning and adaptive behavior, and large-scale AI systems. By grounding artificial systems in these shared principles, IFM opens a pathway toward constructing models whose generalization capabilities arise from neuronal structure rather than model scale alone.
At the same time, IFM highlights critical unresolved questions: \textit{What minimal neuron functions are sufficient to support general-purpose intelligence? Which aspects of biological plasticity are essential for adaptive behavior? How can artificial systems reconcile the tension between biological complexity and computational tractability?} Addressing these questions with IFM could not only advance AGI but also deepen our understanding of intelligence as a natural phenomenon.

Ultimately, IFM provides more than a technical proposal; it outlines a long-term scientific agenda. By integrating biological and functional scaling, IFM charts a stepwise, interpretable, and experimentally grounded path toward general-purpose intelligent systems. Its challenges are substantial, but so too is its potential to reshape how we study, model, and engineer intelligence across biological and artificial domains.

\bibliography{ref}
\bibliographystyle{splncs04}

\end{document}